\documentclass{bmvc2k}

\title{Sketch-based Video Object Segmentation: Benchmark and Analysis}

\addauthor{Ruolin Yang}{yangruolin@bupt.edu.cn}{1}
\addauthor{Da Li}{dali.academic@gmail.com}{2}
\addauthor{Conghui Hu}{conghui@nus.edu.sg}{3}
\addauthor{Timothy Hospedales}{t.hospedales@ed.ac.uk}{2}
\addauthor{Honggang Zhang}{zhhg@bupt.edu.cn}{1}
\addauthor{Yi-Zhe Song}{y.song@surrey.ac.uk}{2}

\addinstitution{
Beijing University of Posts and\\
Telecommunications, Beijing, China
}
\addinstitution{
 SketchX, CVSSP\\
 University of Surrey, UK
}
\addinstitution{
 Department of Computer Science,\\
 National University of Singapore
}

\runninghead{YANG,LI,HU,HOSPEDALES,ZHANG,SONG}{SKETCH-VOS: BENCHMARK AND ANALYSIS}

\usepackage{graphicx}
\usepackage{float} 
\usepackage{subfigure}
\usepackage{amsmath}
\usepackage{amssymb}
\usepackage{booktabs}
\usepackage{colortbl}
\usepackage{xcolor}
\usepackage{makecell}
\usepackage{arydshln}
\usepackage{cellspace}
\usepackage{multirow}
\usepackage[capitalize]{cleveref}
\crefname{section}{Sec.}{Secs.}
\Crefname{section}{Section}{Sections}
\Crefname{table}{Table}{Tables}
\crefname{table}{Tab.}{Tabs.}
\begin{document}

\maketitle

\begin{abstract}
Reference-based video object segmentation is an emerging topic which aims to segment the corresponding target object in each video frame referred by a given reference, such as a language expression or a photo mask. However, language expressions can sometimes be vague in conveying an intended concept and ambiguous when similar objects in one frame are hard to distinguish by language. Meanwhile, photo masks are costly to annotate and less practical to provide in a real application. This paper introduces a new task of sketch-based video object segmentation, an associated benchmark, and a strong baseline. Our benchmark includes three datasets, Sketch-DAVIS16, Sketch-DAVIS17 and Sketch-YouTube-VOS, which exploit human-drawn sketches as an informative yet low-cost reference for video object segmentation. We take advantage of STCN, a popular baseline of semi-supervised VOS task, and evaluate what the most effective design for incorporating a sketch reference is. Experimental results show sketch is more effective yet annotation-efficient than other references, such as photo masks, language and scribble. The datasets are released at \href{https://github.com/YRlin-12/Sketch-VOS-datasets}{https://github.com/YRlin-12/Sketch-VOS-datasets}.
\end{abstract}

\vspace{-2mm}
\section{Introduction}
\label{sec:intro}

\begin{figure*}
\centering
\includegraphics[width=1.0\linewidth]{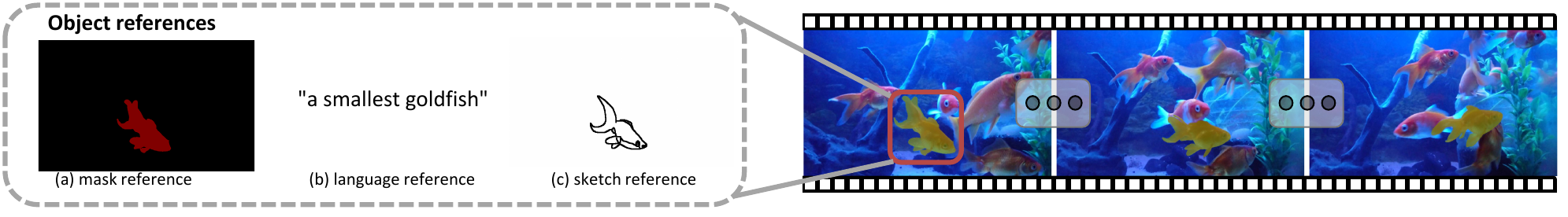}
	\vspace{-0.6cm}
 \caption{A comparison example between three different annotation types for the Semi-VOS task. \textbf{(a)} Mask reference. \textbf{(b)} Language reference. \textbf{(c)} Sketch reference (Ours). }
\label{fig:compare}
	\vspace{-0.7cm} 
\end{figure*}
Video object segmentation (VOS) aims to automatically identify and segment target objects in a given video and has witnessed considerable progress recently. Traditional VOS can be divided into three settings, unsupervised VOS~\cite{pei2022hierarchical, wang2019zero, wang2019learning, yang2019anchor,gowda2020alba}, which segments the most salient object; semi-supervised VOS (Semi-VOS)~\cite{oh2019video, cheng2021rethinking}, which segments the target object given a photo mask reference for the first frame and supervised VOS~\cite{miao2020memory, cheng2021modular, yin2021learning, heo2021guided}, requiring users to interact with the system to refine the output masks repeatedly until the result is satisfactory. However, unsupervised VOS lacks the flexibility to segment an object of interest, while supervised VOS requires intense user interactions. Thus, Semi-VOS is more useful and practical. Semi-VOS methods segment and propagate the object masks given the user annotated masks of the first frame and achieve strong results. However, annotating photo masks for even the first frame only of large-scale datasets like YouTube-VOS~\cite{xu2018youtube} for model training is time-consuming and not practical for users to provide in real applications. To overcome the annotation drawback of photo mask-based semi-VOS, referring VOS has been introduced recently~\cite{khoreva2018video, seo2020urvos}, which employs the language expressions as a new type of reference to guide  object segmentation in VOS. However, despite its general efficacy, it is sometimes challenging to convey some concept effectively with words. 

In recent years, sketch, as a complementary modality to text, has been investigated broadly due to the demand for interacting tools on popular touchscreen devices. The common view is that sketch is preferable when words are inconvenient to convey an intended concept. At the same time, the fact that the time necessary to create one sketch (54.84 secs) is less than that for one photo segmentation mask (109.01 secs)~\cite{hu2020sketch} demonstrates that drawing sketches is a more time-efficient user annotation. 
Sketches contain category and fine-grained level information verified by various sketch-based image retrieval works \cite{dey2019doodle, collomosse2019livesketch, sain2021stylemeup,chowdhury2022partially,bhunia2020sketch,bhunia2021more,bhunia2022sketching}. While sketch-based image editing works \cite{zeng2022sketchedit, wang2021sketch} also demonstrate the expressiveness and editability of sketches. Sketches can be used as a query to classify, segment or locate novel objects in an input image, as shown in Sketch-a-Classifier~\cite{hu2018sketch}, Sketch-a-Segmenter~\cite{hu2020sketch} and sketch guided localization (SGL)~\cite{tripathi2020sketch}, respectively. However, the existing sketch-based benchmarks have only considered sketches for image-level tasks, giving less consideration to the video domain.

In this paper, we propose sketch-based video object segmentation and introduce a new task, Sketch-VOS, to predict photo masks in video frames by sketch references given for the first frames. Our three Sketch-VOS datasets are extended from DAVIS16~\cite{perazzi2016benchmark}, DAVIS17~\cite{pont20172017} and YouTube-VOS~\cite{xu2018youtube} datasets with free-hand sketches following the rules of Ref-DAVIS~\cite{khoreva2018video} and Ref-YouTube~\cite{seo2020urvos} for fair comparisons to the existing works. In Figure \ref{fig:compare}, we provide a comparison example between three different types of references, photo mask, language expression and sketch. As shown in Figure \ref{fig:compare} (b), language expressions may be ambiguous in situations where objects are similar. Whereas fine-grained information like the shape and pose of a drawn sketch can easily overcome this limitation. As the first dataset that pairs sketches with video objects, our dataset will allow researchers to develop more annotation-efficient algorithms for video object segmentation and potentially other associated problems like video editing.

Our paper is structured to address three questions: 
(i) What is sketch-based video object segmentation? We will detail how we construct our three datasets for sketch-based VOS and how to use them for model evaluation. 
(iii) Why is sketch a better reference for VOS than photo mask, language, and scribble? We will compare our sketch based STCN with other SOTA VOS methods using other references.

\vspace{-5mm}
\section{Related Work}
\label{sec:related}
\begin{figure}[!t]
	\centering
	\begin{minipage}{0.65\linewidth}
		\centering
		\includegraphics[width=1.0\linewidth]{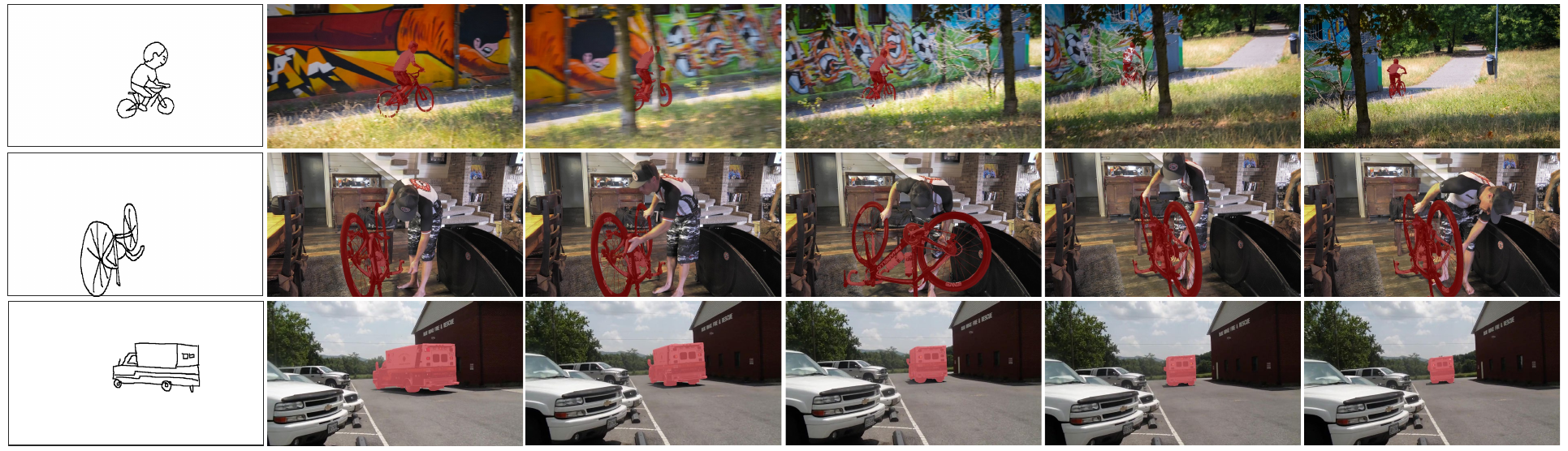}
	\end{minipage}
 \hspace{3mm}
	\begin{minipage}{0.3\linewidth}
		\centering
		\includegraphics[width=1.0\linewidth]{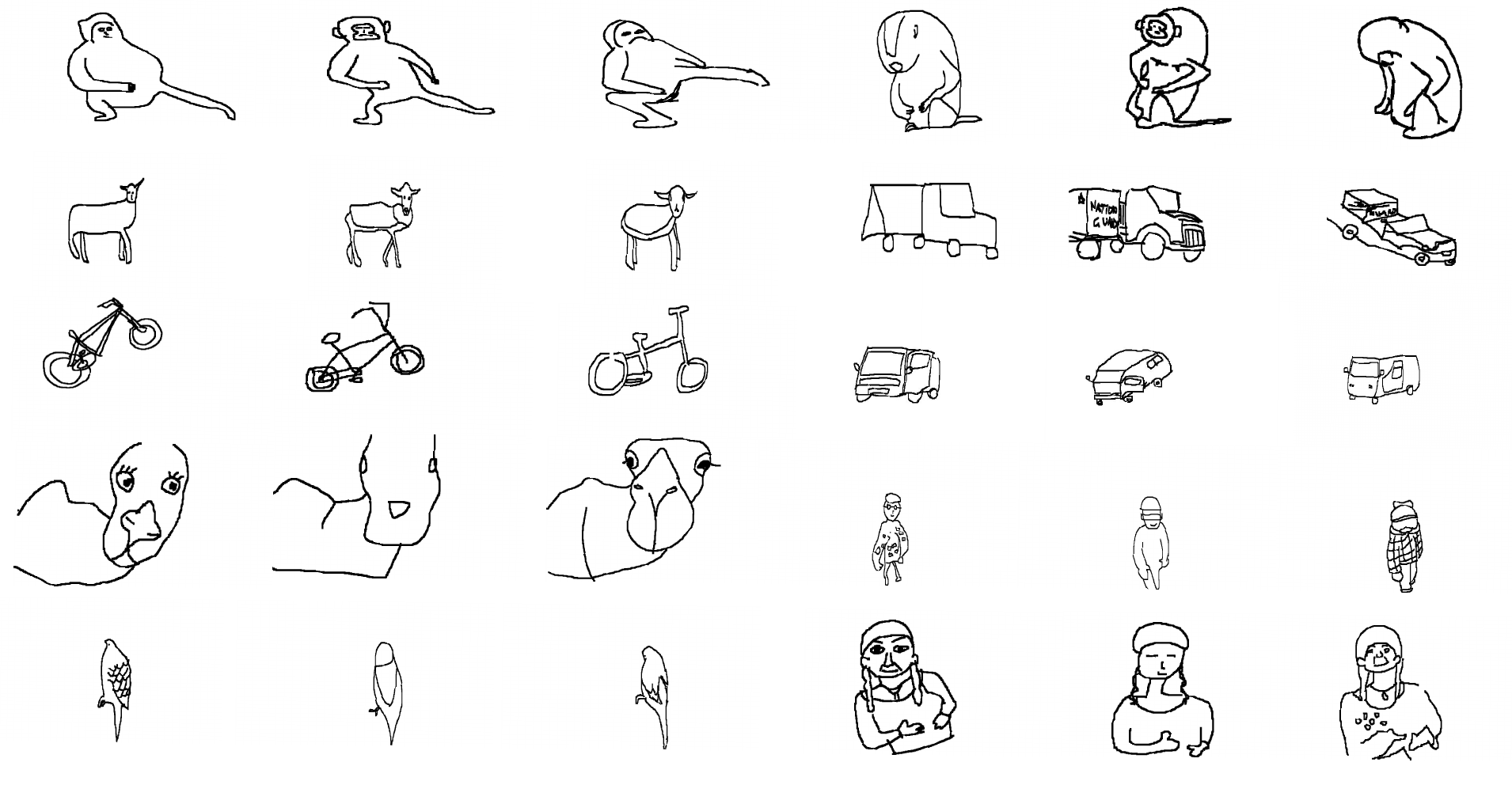}
	\end{minipage}
 \caption{Left: Sketch reference examples of Sketch-DAVIS16 (row 1), Sketch-DAVIS17 (row 2) and Sketch-Youtube-VOS dataset (row 3). Right: examples of references in our Sketch-VOS benchmark.}
		\label{fig:diversity}
 \vspace{-5mm}
\end{figure}
\paragraph{(Photo Mask-based) Semi-supervised Video Object Segmentation.} 
Early semi-supervised video object segmentation focused on fine-tuning at test time~\cite{caelles2017one, cheng2018fast, cheng2017segflow, liu2022learning} or matching and propagating~\cite{yang2021collaborative, cho2022tackling, mao2021joint, voigtlaender2019feelvos,oh2019video,cheng2021rethinking} the pixel-level object(s) mask(s) of the first frame. The latter one is more efficient and usually trained end to end. STM~\cite{oh2019video} made remarkable progress by using space-time memory bank and became the backbone of many following state-of-art methods~\cite{cheng2021rethinking, lin2022swem, liu2022learning, liu2022global}. STCN~\cite{cheng2021rethinking} improved the affinity and feature extraction of STM and reached a more effective and efficient version. In this work, we focus on modifying and extending STCN for experimentation and analysis due to its robustness of temporal coherence. 

\vspace{-5mm}
\paragraph{(Language-based) Referring Video Object Segmentation.}
Recently, referring video object segmentation (RVOS) has attracted great attention from researchers. This task aims to segment and track the mask of the target object in a video referred by a language expression. \cite{khoreva2018video} released Ref-DAVIS dataset at first place with 90 videos and employed a complicated model which located object by bounding box first and then propagated to predict the mask. Then, URVOS~\cite{seo2020urvos} provided a large-scale referring video object segmentation dataset (Ref-YouTube-VOS) and introduced an STM-style model with cross-modal attention block to fuse frame feature and text feature. 
ReferFormer~\cite{wu2022language} achieved state-of-the-art results of RVOS by a top-down method with the support of the popular detector Deformable-DETR~\cite{zhu2020deformable}. Language expression as the Transformer decoder input is the key component of this approach and Hungarian matching~\cite{kuhn1955hungarian} is required for linking the instance tube. However, it relies on high computation resources due to its complexity.

\vspace{-5mm}
\paragraph{Sketch as Queries.} 
Sketch-based image retrieval is a fundamental task that aims to retrieve photos of the same category~\cite{dey2019doodle, collomosse2019livesketch, ribeiro2020sketchformer} or corresponding instance~\cite{sain2021stylemeup,chowdhury2022partially,bhunia2020sketch,bhunia2021more,bhunia2022sketching} given a query sketch. Sketch-a-Classifier~\cite{hu2018sketch} designed a model to generate a photo classifier by giving a sketch of an unseen category. Sketch-a-Segmenter~\cite{hu2020sketch} similarly were made to produce a novel pixel-level classifier by a sketch input. SGL~\cite{tripathi2020sketch} proposed to generate object proposals relevant to the sketch query. DIY~\cite{bhunia2022doodle} employed sketch queries to achieve the goal of few-shot class incremental learning. These methods mainly focused on image-level tasks and paid less attention to video-sketch correspondence. To the best of our knowledge, this is the first work to apply sketch to the video object segmentation task.

\vspace{-5mm}
\section{Sketch-based VOS Benchmark}
\label{sec:dataset}
We extend three popular VOS benchmarks including DAVIS16~\cite{perazzi2016benchmark}, DAVIS17\cite{pont20172017} and YouTube-VOS~\cite{xu2018youtube} with first-frame sketch annotations for segmenting target objects in video sequences. Examples are illustrated in Figure \ref{fig:diversity} (left). A detailed comparison of the datasets is given in the supplementary file.

\vspace{-5mm}
\paragraph{Data Collection and Pre-Processing}
The sketch data is collected by a collection interface following FSCOCO~\cite{chowdhury2022fs} dataset. Similar to the language reference~\cite{khoreva2018video}, we asked the participants to sketch for the target objects appearing in the first frame without seeing the full video. We provided each participant with a randomly selected object, as well as a blank canvas to sketch on. The participants had 60 seconds to remember as many details as they could about the pose, shape, and fine-grained characteristics, before the object is removed. The intention is to create a sketch that will make it possible for those who have never seen the video to identify the target object. To verify the quality of our sketch, we requested assistance from 20 volunteers to validate our dataset. Each of them would be provided a video and a sketch corresponding to one object. At the beginning, the first frame would last for a while and the video would play, volunteers then used the bounding box to label the object in the video. The final step is to re-draw references for cases where manual VOS above failed. To keep the diversity of sketch, we did not train any participants and asked them to draw in their own style. As shown in Figure \ref{fig:diversity} (right), the participants all drew the sketch in different styles, but salient visual properties (e.g., pose) of each object were uniformly depicted. On average each object has been annotated with three sketches and it takes the annotator around 30s to draw for a target object. One big challenge of VOS is there are many similar-looking instances in one video as shown in Figure \ref{fig:compare}. And according to Sketch-a-Segmenter~\cite{hu2020sketch}, sketches with position and scale alignment will benefit segmenting instance-level objects. Therefore, we subject all sketches to this preprocessing strategy before our experimental evaluation.

\vspace{-5mm}
\paragraph{Sketch-DAVIS-VOS}
DAVIS16~\cite{perazzi2016benchmark} is a dataset where only a single object is annotated per frame, which comprised of 30 training and 20 validation videos from four evenly dispersed classes (humans, animals, vehicles, objects) with all the frames annotated with pixel-level accuracy. Then DAVIS17~\cite{pont20172017} extended DAVIS16 to 60 training and 30 validation videos and annotated multiple objects. Ref-DAVIS~\cite{khoreva2018video} extended DAVIS datasets with two referring expression based on first frame only as well as full video. The latter one is different from mask annotation where the annotator will describe the object after viewing the full video. We only provide first frame annotations same as the mask annotations. To the end, we collect 150 sketches for DAVIS16, and 615 sketches for DAVIS17.

\vspace{-5mm}
\paragraph{Sketch-YouTube-VOS}
YouTube-VOS\cite{xu2018youtube} is a large-scale multi-object VOS benchmark consisting of 3417 training videos of 65 categories and 507 validation videos of 65 training classes and 26 unseen categories. Ref-YouTube-VOS dataset~\cite{seo2020urvos} contains the language expressions for two type -- first frame and full video. However, some objects can not be identified by words. Therefore, Ref-YouTube-VOS only takes partial objects from YouTube-VOS to annotate. Though sketch can be used to refer any objects in frames, we only draw sketches for objects appeared in Ref-YouTube-VOS for fair comparison. Note that only the ground-truth masks of YouTube-VOS training set is publicly available. The validation set results can be only evaluated on the competition server. And only the evaluation entry for the full-video language expressions is currently open on the server. Therefore, we draw sketches for the training set based on the language expressions of first frame and validation set based on the expressions of full video.

\vspace{-5mm}
\section{Sketch-based VOS Model}
\label{sec:experiments1}

This section provides an overview of how sketch can be used as a new type of reference in a reference-based VOS model. We employ the popular Semi-VOS method STCN \cite{cheng2021rethinking} as our baseline and explore the interaction between two modalities -- sketch and video. We extend STCN with various fusion designs, such as input fusion, latent fusion and sketch-based weight generation.
\begin{figure*}[t]
\begin{center}
   \includegraphics[width=1.0\linewidth]{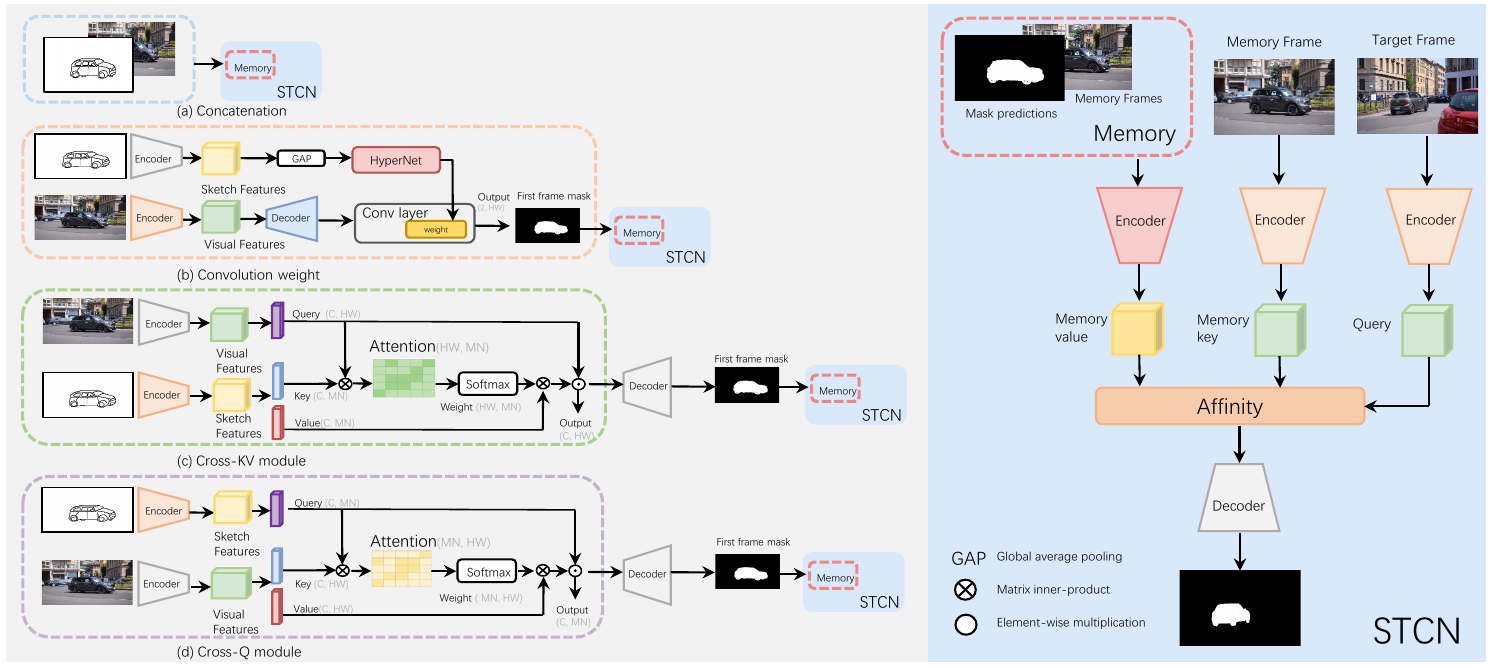}
\end{center}
\vspace{-4mm}
   \caption{ The Sketch-based VOS model with various designs: (a) Concatenation, (b) Convolution weight, (c)  Cross-KV, and (d) Cross-Q.}
\label{fig:method}
\vspace{-5mm}
\end{figure*}

\vspace{-2.5mm}
\subsection{STCN} 
We first explain how STCN works before introducing our model. As shown in the right part of Figure~\ref{fig:method}, STCN is a memory-based method which encodes frames to keys and queries and encodes masks (concatenated with frames) to values. Every time the feature of a target frame is updated by an affinity between query features and memory key/value features in the memory bank. Then, features are gradually processed and upsampled by the decoder.

\vspace{-2.5mm}
\subsection{Design-Space of Sketch-based VOS}\label{subsec:designs}
\paragraph{Concatenation} The simplest way to combine sketch and video frames is by concatenating them at the input level. As shown in Figure \ref{fig:method} (a), we encode sketch and frame together as the memory value instead of mask in STCN.

\vspace{-5mm}
\paragraph{Convolution Weight:}~Inspired by Sketch-a-Segmenter\cite{hu2020sketch}, we use a  HyperNet\cite{ha2016hypernetworks} to generate an  instance-level weight for the video segmentation head to predict object mask in the first frames as shown in Figure \ref{fig:method} (b). Then we store the prediction as memory and propagate it through the video. More clearly, Sketch-a-Segmenter predicts masks for all instances given a sketch, while our task is to segment one particular instance from all at a time.

\vspace{-5mm}
\paragraph{Cross Attention}
\label{subsec:cross-model}
Given the first frame of a video input and a sketch reference, their features are generated by a visual encoder and a sketch encoder separately as follows: $\mathcal{F} \in \mathbb{R}^{C_v\times H \times W}$, $\mathcal{S}\in \mathbb{R}^{C_s\times M \times N}$, where $H$, $W$, $M$, $N$ are the spatial dimensions, $C_v$ and $C_s$ are the channel dimensions. We construct two cross-modal attention design strategies to fuse the features. Figure \ref{fig:method} (c) illustrates our first attention module, motivated by ReferFormer~\cite{wu2022language} and LAVT~\cite{yang2022lavt}, we encode sketch as {K}ey and {V}alue by two $1\times1$ convolution mappings. And to match the dimension of sketch feature, we encode visual feature as {Q}uery by another $1\times1$ convolution filter. The outputs are Key $\mathcal{K}_{s} \in \mathbb{R}^{C\times MN}$, Value $\mathcal{V}_{s} \in \mathbb{R}^{C\times MN}$ and Query $\mathcal{Q}_{f} \in \mathbb{R}^{C\times HW}$. Then, a dot-product attention~\cite{vaswani2017attention} is computed between Query and Key. The attention map stores the correspondences between the visual feature and the sketch reference. Then Value will be transformed by the attention map. We called this module {Cross-KV}. The output features $\mathcal{O} \in \mathbb{R}^{C\times HW}$ can be obtained as follows:
\begin{align}   
    \label{eq:1}
    \text{W}= \text{Softmax}(\frac{{Q_f}^{\text{T}}{K_s}}{\sqrt{C}}), \\ \label{eq:2}
    \text{O}= {V_s}W^{\text{T}}\odot {Q_f},
\end{align}
where $\odot$ denotes element-wise multiplication, which was introduced by~\cite{yang2022lavt} and can be explored with other options. We also provide a novel cross-modal attention module as shown in Figure~\ref{fig:method} (d). Since sketch is the reference condition that leads a model to segment a target mask, we use a sketch feature as the Query in cross-modal attention. Similar to the above, we gain the new Key $\mathcal{K}_{f} \in \mathbb{R}^{C\times HW}$, Value $\mathcal{V}_{f} \in \mathbb{R}^{C\times HW}$ and Query $\mathcal{Q}_{s} \in \mathbb{R}^{C\times MN}$. This time we compute the attention map between Query from the sketch feature and Key from the visual feature. This Attention map is simply the transposed matrix of the attention map mentioned above, but the Value now is from the visual feature.
We called this design {Cross-Q}. The output features $\mathcal{O} \in \mathbb{R}^{C\times MN}$ can be obtained as follows:
\begin{align}   
    \label{eq:3}
    \text{W}= \text{Softmax}(\frac{{Q_s}^{\text{T}}{K_f}}{\sqrt{C}}), \\ \label{eq:4}
    \text{O}= {V_f}W^{\text{T}}\odot {Q_s},
\end{align}

The output features are fed into the STCN decoder which generates a binary mask of the first frame and then propagates this mask by the memory bank to segment the remaining target object masks in the video.

\vspace{-5mm}
\section{Experiments}
\label{sec:experiments1}

We evaluate the performance of our Sketch-based VOS model and compare sketch with other references, such as photo mask, language expression and scribble.

\vspace{-3mm}
\subsection{Experimental Setup}
\noindent\textbf{Evaluation Metrics:} We evaluate our reuslts by the standard evaluation metrics~\cite{perazzi2016benchmark} for VOS tasks, i.e., region similarity $\mathcal{J}$, contour accuracy $\mathcal{F}$, and the average of $\mathcal{J}$ and $\mathcal{F}$ ($\mathcal{J} \& \mathcal{F}$). For DAVIS dataset, we evaluate by the official evaluation code \footnote{\textcolor{magenta}{https://github.com/davisvideochallenge/davis2017-evaluation}}. All experiments on YouTube datasets are evaluated on the competition server \footnote{\textcolor{magenta}{https://youtube-vos.org/dataset/rvos/}} same as Ref-YouTube-VOS method~\cite{wu2022language}. Note that only 202 videos in validation set can be evaluated on the server. All the following results are based on these 202 videos.

\noindent\textbf{Implementation Details:} Following STCN~\cite{cheng2021rethinking}, every video frame and corresponding sketch are downscaled to 384p. We train our model using the Adam optimizer with initial learning rate of 1e-5. The frame encoder is initialized with classification weights pre-trained on ImageNet\cite{deng2009imagenet} while sketch encoder is initialized with classification weights pre-trained on QuickDraw~\cite{ha2017neural}. Different from STCN, we pick first frame as a default frame and randomly sample other two temporal frames to form a training clip. We use BCE Loss for all experiments.

\vspace{-3mm}
\subsection{Sketch-based VOS models Results}
We encode sketches and video frames by two separate ResNet50~\cite{he2016resnet}. We first conduct the ablation study of various designs proposed in~\ref{subsec:designs}.

For {Cross-KV} and {Cross-Q}, we also tried different levels of features and different variants of interacting sketch reference and video frames as follows:
\emph{GAP sketch features} -- {Given the output from layer4 of the sketch encoder, we aggregate global information of a sketch using global average pooling (GAP). This gives a reference to verify the importance of the spatial information embedded in sketch features. }
\emph{Spatial sketch features} -- Namely, we use the spatial feature maps generated from the sketch encoder, which retains the spatial information.
 \emph{Multi-level visual features} -- i.e. features from multiple layers of a video encoder.

 \begin{table}[!t]
\centering
\centering
\caption{Sketch-based VOS Model evaluated on Sketch-YouTube-VOS validation set. GAP indicates global average pooling. Decoder means the convolution layer of segmentation head of Decoder. HyperNet means the weight generated by HyperNet.}
   \label{tbl:fusion}
 \resizebox{\columnwidth}{!}{
\begin{tabular}{l c c c c c cc c c}
        \toprule
      \multirow{2}{*}{Fusion Designs}
      &\multicolumn{2}{c}{Level} 
      & \multirow{2}{*}{\textbf{$\mathcal{J} \& \mathcal{F} $}} 
      & \multirow{2}{*}{\textbf{$\mathcal{J}$}}
      &\multirow{2}{*}{\textbf{$\mathcal{F}$}}  
      & \multirow{2}{*}{ }
      &\multirow{2}{*}{\textbf{$\mathcal{J} \& \mathcal{F} $} }
      & \multirow{2}{*}{\textbf{$\mathcal{J}$}}
      & \multirow{2}{*}{\textbf{$\mathcal{F}$} }\\
      \cline{2-3}
       & visual &sketch \\
       \hline
       
        \textbf{(a) Concatenation}
         &raw pixel&raw pixel &74.1&72.1 &76.0  & &-  &- &-  \\
         \hline  

        {\textbf{(b) Convolution weight}}
         &Decoder&HyperNet   &19.6 &20.1 &19.0 & &- &- &-  \\
         \hline 
         \multirow{6}{*}{{\textbf{(c) Cross-KV/Q}}} & & &\multicolumn{3}{c}{Cross-KV} & &\multicolumn{3}{c}{Cross-Q} \\
          \cdashline{4-6}   \cdashline{8-10}
         &res5&GAP   &19.6  &16.2  &23.1&  &55.1  &54.1  &56.1\\
         &res4&res4 &56.6   &55.5  &57.6 &  &74.3  &72.3  &76.3\\
         &res5&res5 &67.1  &65.4  &68.7&   &74.8  &72.8  &76.8\\
         &multi-level&res4&57.3  &55.9 &58.6 &  &74.7  &72.7 &76.7\\
          &multi-level&res5 &55.8 &54.4 &57.1 &  &\textbf{74.9}  &\textbf{72.9} &\textbf{77.0}\\
         \bottomrule
         
    \end{tabular} }

 \vspace{-0.3cm}  	
\end{table}

 Table \ref{tbl:fusion} reports the results for various sketch-based VOS models. We can see that designs using cross attention and input concatenation all give reasonable results and the best one boosts the performance to \textbf{$\mathcal{J} \& \mathcal{F} $} of $74.9$, $\mathcal{J}$ of $72.9$ and $\mathcal{F}$ of $77.0$, whereas sketch-based weights generation does not work in this task. This may be the reason that, in this case, sketch reference is interacting less with video frames and the reference information is hard to propagate among video frames. Among cross-attention variants, Cross-Q works much better than Cross-KV and improves the \textbf{$\mathcal{J} \& \mathcal{F} $} by more than around $7$ absolute points, which is interestingly different from the existing wisdom from~\cite{wu2022language, yang2022lavt}, indicating sketch as a reference may require different designs than language. Without a surprise, using GAP features works badly and indicates the spatial information provided by sketch is crucial for the VOS task. Using multi-level visual features improves the performance marginally, which may not be favoured considering the extra computational cost it brings.

\begin{figure}[!t]
\begin{center}
   \includegraphics[width=1.0\linewidth]{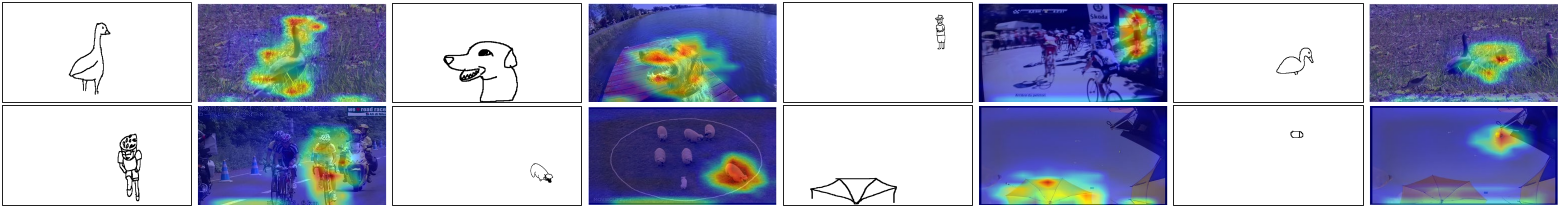}
\end{center}
\vspace{-4mm}
   \caption{ Visualized sketch queries and corresponding feature maps weighted by attention on Sketch-YouTube-VOS validation set.}
\label{fig:heatmap}
\vspace{-4mm}
\end{figure}

We also visualise the generated attention maps by querying the first frame from various videos of Sketch-YouTube-VOS validation set using sketch references. As illustrated in Figure \ref{fig:heatmap}, we can see that the VOS model can attend to the indicated areas by the sketch references precisely regardless the categories and scales of the target objects. Even an object which is too small to be described by language can be localized precisely by sketch.

In summary, using sketch features as Query in the cross-modal attention with a spatial sketch feature works effectively on the video object segmentation task.

\begin{table*}[t]
\centering
\caption{Comparison with state-of-the-art methods on Youtube-VOS, DAVIS17 and DAVIS16 datasets.}
 \label{tbl:ytb}
 \resizebox{\columnwidth}{!}{
\begin{tabular}{cl ccc ccc ccc}
        \toprule
       \multirow{2}{*}{Reference} &\multirow{2}{*}{Method} &\multicolumn{3}{c}{Youtube-VOS} &\multicolumn{3}{c}{DAVIS17} &\multicolumn{3}{c}{DAVIS16}\\
       & &\textbf{$\mathcal{J} \& \mathcal{F}$} & \textbf{$\mathcal{J}$} & \textbf{$\mathcal{F}$}
    &\textbf{$\mathcal{J} \& \mathcal{F} $} & \textbf{$\mathcal{J}$} & \textbf{$\mathcal{F}$}
        &\textbf{$\mathcal{J} \& \mathcal{F} $} & \textbf{$\mathcal{J}$} & \textbf{$\mathcal{F}$}\\
       \hline
        \multirow{6}{*}{\textbf{Text}}
         &VOSwL\cite{khoreva2018video}&- &- &- &39.3 &37.3 &41.3 &84.1 &82.8 &85.4 \\
         &URVOS\cite{seo2020urvos} &46.5 &44.2 &48.8 &51.7 &47.3 &56.0  &- &- &- \\
        &HINet\cite{yang2021hierarchical} &- &- &- &52.0 &- &- &84.8 &84.4 &85.3 \\
        &YOFO\cite{li2022you} &48.6 &47.5 &50.0&55.4 &50.1 &58.7 &- &- &-\\
        &MLRL\cite{wu2022multi} &49.7 &48.4 &51.0&57.9 &53.9 &62.0 &- &- &- \\
        &LBDT\cite{ding2022language} &49.4 &48.2 &50.6 &54.1 &- &- &- &- &- \\
        &MTTR\cite{botach2022end} &55.3 &54.0 &56.6 &- &- &- &- &- &- \\
        &ReferFormer\cite{wu2022language}  &64.9 &62.8 &67.0 &61.1 &58.1 &64.1 &- &- &- \\
        \hline
        \multirow{2}{*}{\textbf{Mask}}
         &STM\cite{oh2019video} &74.7 &72.8 &76.6&69.5 & 67.0 &72.0  &- &- &- \\
        &STCN\cite{cheng2021rethinking} &79.6  &77.1  &82.1&74.4  &71.5 &77.2 &- &- &- \\
        \hline
        \rowcolor[rgb]{ .94,  .94,  .94}
        \textbf{Sketch} &Ours   &75.4 &73.4 &77.5 &70.2 &66.9 &73.4 &81.6 &80.2 &83.1  \\
         \bottomrule
    \end{tabular}
    }
 
\vspace{-0.3cm}
\end{table*}

\vspace{-0.3cm}
\subsection{Sketch v.s. Other References}

We compare our Sketch-VOS with the existing VOS works incorporated with different references that appeared in the literature.Furthermore, we conduct comprehensive comparisons between several references: language, sketch, mask, and interactive annotations including cross, circle, scribble and object contour.
\vspace{-0.5cm}
\paragraph{Baselines:} (i) State-of-the-art methods: VOSwL~\cite{khoreva2018video} predicts masks by localising and segmenting in two stages. URVOS~\cite{seo2020urvos} is similar to our design but they concatenate visual and linguistic features before feeding into the cross-modal attention. YOFO~\cite{li2022you} transfers object information by meta-learning. HINet\cite{yang2021hierarchical} employs a hierarchical fusion of language and frame. MLRL~\cite{wu2022multi} fuses linguistic features with video, frame and object features in different levels. LBDT~\cite{ding2022language} transfers spatial and temporal visual features by language. MTTR~\cite{botach2022end} and ReferFormer~\cite{wu2022language} use transformer-based detectors to localise and segment masks. STM~\cite{oh2019video} designs a spatial-temporal memory bank, then STCN~\cite{cheng2021rethinking} improves it with a more effective affinity module. (ii) Fair comparison baselines: All experiments use STCN as the backbone and follow the same implementation setting in \cref{sec:experiments1}. The fusion methods vary in different modalities. Since YouTube-VOS dataset does not collect scribbles as annotations, we extend YouTube-VOS dataset with scribbles in the first frame by following \cite{caelles20182018}. We simply concatenate it with the first frame before feeding it into STCN following~\cite{cheng2021modular}. As for language expression, we encode first-frame expressions by the popular language encoder BERT~\cite{vaswani2017attention} initialized by the official pre-trained weights. We utilize {Cross-KV} module to fuse the linguistic features and frame figures following language-based methods~\cite{yang2022lavt,wu2022language}. We did not pre-train the mask-based STCN on extra static image datasets for a fair comparison. (iii)Ablation Study of different references: To verify the effectiveness of sketch, we further compared sketch with text+click, text+bounding box, cross, circle and contour. We extend YouTube-VOS dataset in the first frame with theses interactive annotations. Specifically, the cross and the click are drawn on the center point of the ground-truth mask; the circle and the box are obtained by fitting an outer circle/box to the ground-truth mask; the contour is acquired by computing the convex hull of the sketch. All experiments use STCN as the backbone and follow the same implementation setting in \cref{sec:experiments1}.

\begin{table}[!th]
\centering
\begin{minipage}[t]{0.48\textwidth}
\centering
\caption{Fair comparison with different references on YouTube-VOS validation dataset. The rightmost column shows the average time of annotating one object.}
    \label{tbl:fair}
    \scalebox{0.8}{
\begin{tabular}{l c c c c}
        \toprule
       Reference & \textbf{$\mathcal{J} \& \mathcal{F} $} & \textbf{$\mathcal{J}$} & \textbf{$\mathcal{F}$} & Annotating time\\\hline
        \textbf{Text} &44.4  &42.7  &46.1 &5.0s\cite{khoreva2018video} \\
         \textbf{Scribble} &69.1 &67.3 &71.0&1.25s \cite{lin2016scribblesup}\\ 
        \rowcolor[rgb]{ .94,  .94,  .94}
         \textbf{Sketch}&75.4 &73.4 &77.5 &30.6s\\
        \textbf{Mask}&79.6  &77.1  &82.1 &109.0s\cite{hu2020sketch} \\
         \bottomrule      
    \end{tabular}}

\end{minipage}
\hspace{3mm}
\begin{minipage}[t]{0.48\textwidth}
\centering
\caption{Ablation study of different references on YouTube-VOS validation dataset.}
    \label{tbl:fair_alb}
    \scalebox{0.8}{
\begin{tabular}{l c c c}
        \toprule
       Reference & \textbf{$\mathcal{J} \& \mathcal{F} $} & \textbf{$\mathcal{J}$} & \textbf{$\mathcal{F}$} \\\hline
       Text &44.4  &42.7  &46.1\\
       Text+Click &57.0 &54.4 &59.6 \\
        Text+Box  &58.0  &55.3  &60.6 \\ 
        \hline
         Cross&56.1  &53.2  &59.1  \\
        Circle&58.6  &56.6  &60.6 \\
       Contour&71.8  &69.6  &74.1 \\
       \hline
        \rowcolor[rgb]{ .94,  .94,  .94}
         \textbf{Sketch}&75.4 &73.4 &77.5 \\
         \bottomrule      
    \end{tabular}}
\end{minipage}
\vspace{-0.3cm}
\end{table}

\vspace{-5mm}
\paragraph{YouTube-VOS.}~Table \ref{tbl:ytb} shows the results of the recent state-of-the-art video segmentation methods on the YouTube-VOS validation set. Our best model achieves a competitive \textbf{$\mathcal{J} \& \mathcal{F} $} among all competitors.  Nevertheless, our model outperforms all language-based methods under all metrics and by significant margins.
For a fair comparison with mask-based methods, we retrain the STM~\cite{oh2019video} and STCN~\cite{cheng2021rethinking} without any extended image datasets. Our model demonstrates better performance compared to STM, albeit slightly trailing STCN.

\vspace{-5mm}
\paragraph{DAVIS.}~In Table \ref{tbl:ytb}, we evaluate our model on the DAVIS17 validation set. Due to its small scale, we directly evaluate this dataset using models trained on YouTube-VOS. However, DAVIS has longer videos and is annotated more strictly than YouTube-VOS which is more challenging for VOS. The results show that our model can easily generalize to another dataset, and again our method outperforms all language-based methods without using extra image datasets. Compared to the mask-based methods, we outperform STM by $0.7$ points and underperform STCN by $4.2$ points in terms of \textbf{$\mathcal{J} \& \mathcal{F} $} metrics. Table \ref{tbl:ytb} shows the results on DAVIS16. Competitors are limited in this case, including only VOSwL \cite{khoreva2018video} and HINet \cite{yang2021hierarchical}, nevertheless our sketch-VOS has a close performance to these two mask-based VOS methods.
Please see more fine-tuning results on DAVIS datasets in the supplementary.

\begin{figure}[!th]
\centering
\includegraphics[width=1.0\linewidth]{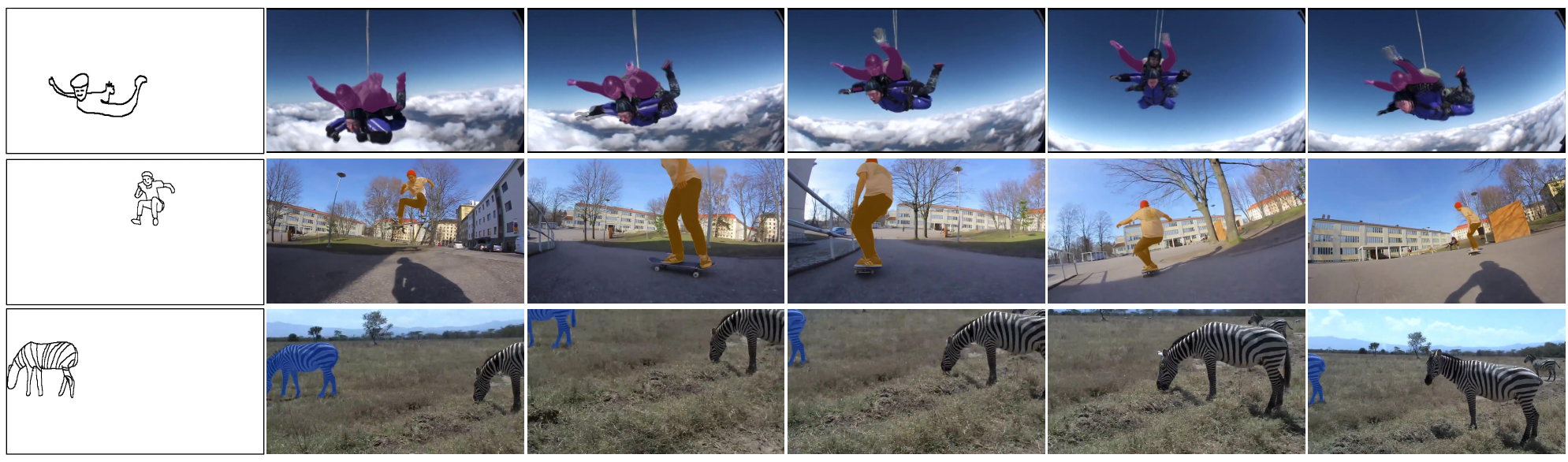}
	\vspace{-0.5cm}
 \caption{Qualitative results on the Sketch-YouTube-VOS validation set. Best viewed in color.}
\label{fig:example}
	\vspace{-0.3cm}
\end{figure}

\vspace{-5mm}
 \paragraph{Visualizations.}~Figure \ref{fig:example} visualizes results on Sketch-YouTube-VOS dataset. Our model can successfully segment the object mask in each frame given a sketch reference. Our model is robust even in situations with multiple similar objects, appearance changing, fast motion and outside the frame. 
 More visualization of our sketch-VOS results on Sketch-DAVIS datasets can be found in the supplementary. 
 In Figure \ref{fig:example-compare}, we visualize the comparison between the state-of-art language referring VOS method ReferFormer and our method. 
 ReferFormer loses track of the target object as there are many similar objects in the first video. However, our model can track and segment the referred duck consistently. 
 In the more difficult second video, ReferFormer can not distinguish the target object at all, while our sketch-VOS conducts the perfect segmentation due to the embedded distinctive fine-grained information.

\begin{figure}[!th]
\centering
\includegraphics[width=1.0\linewidth]{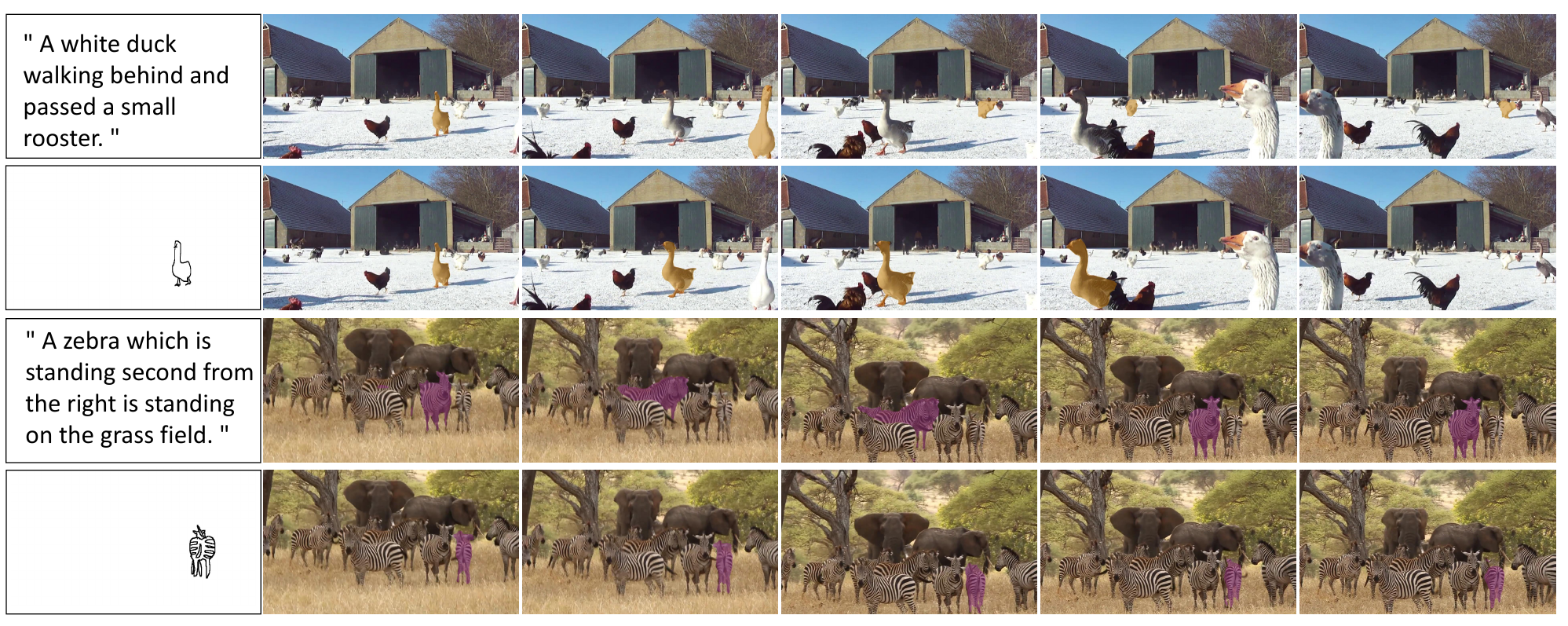}
	\vspace{-0.5cm}
 \caption{Visual comparison with ReferFormer \cite{wu2022language} on the YouTube-VOS validation set.  Best viewed in color.}
\label{fig:example-compare}
	\vspace{-0.4cm}
\end{figure}

\vspace{-5mm}
\paragraph{Fair comparison between different references.}
Table \ref{tbl:fair} shows the fair comparison of different references.
We can see in such case, sketch outperforms text by more than $20$ \textbf{$\mathcal{J} \& \mathcal{F} $}  points. Scribble also works well in this setting but still does not achieve comparable performance to sketch, with a margin of $6.3$ \textbf{$\mathcal{J} \& \mathcal{F} $}. Without pre-training on huge static image datasets, mask reference only achieves 79.6 of \textbf{$\mathcal{J} \& \mathcal{F} $}. All these results suggest that sketch can be effectively used as an alternative and cheaper reference for video object segmentation. 
Additional visual comparisons are provided in the supplementary.
\vspace{-5mm}
\paragraph{Ablation Study of different references}From the results in Tab.~\ref{tbl:fair_alb}, sketch can brings more benefits than simple indicator like cross, circle, box or click. This is because they cannot provide additional information beyond object location, such as semantic context or pose. We can also see that only keeping the contour performs worse than the whole sketch. Our speculation is that the inclusion of fine-grained details in the sketch aids in effectively representing and segmenting the object, whereas relying solely on contours may cause confusion in subsequent video frames.

\vspace{-5mm}
\section{Limitations and Future Directions}
\label{sec:limitation}
Despite the fact that Sketch-VOS datasets are the largest public datasets for sketch-based video object segmentation to date, they are still smaller than the standard large-scale benchmarks. However, a sketch is much cheaper to collect than a photo mask and can be manipulated with relative ease to generate variants for data augmentation ~\cite{yu2015sketch, yu2016sketch}. 

In the future, we will investigate ways for combining motion information with sketches to refer to dynamic object activity. We also plan to increase dataset diversity by generative models and other data augmentation techniques.
\vspace{-5mm}
\section{Conclusion}
\label{sec:conclusion}
We introduced three instance-level datasets for sketch-based video object segmentation. We evaluate our datasets by extending the popular VOS method STCN and explore various fusion designs for better aggregating sketch and visual features. The experimental results show that our method can easily beat language-referring VOS methods and is comparable to mask-based VOS methods. We hope our proposed datasets will drive future research in the field and inspire people to see the potential of sketch for solving complex video tasks.
\bibliography{egbib}
\end{document}